\pdfoutput=1
\documentclass[10pt,twocolumn,letterpaper]{article}

\usepackage{iccv}
\usepackage{times}
\usepackage{epsfig}
\usepackage{graphicx}
\usepackage{amsmath}
\usepackage{amssymb}

\usepackage[title]{appendix}
\usepackage{algorithm}
\usepackage{algorithmicx}
\usepackage{algpseudocode}
\usepackage{array}  

\algdef{SE}[DOWHILE]{Do}{doWhile}{\algorithmicdo}[1]{\algorithmicwhile\ #1}%

\renewcommand{\thefootnote}{\fnsymbol{footnote}}

\usepackage[breaklinks=true,bookmarks=false]{hyperref}

\iccvfinalcopy 

\def\iccvPaperID{****} 

\ificcvfinal\fi

\begin{document}

\title{Ensemble-in-One: Learning Ensemble within Random Gated Networks for Enhanced Adversarial Robustness}

\author{Yi Cai\\
Dept. of E.E.\\
Tsinghua University\\
{\tt\small caiy17@mails.tsinghua.edu.cn}
\and
Xuefei Ning\\
Dept. of E.E.\\
Tsinghua University\\
{\tt\small foxdoraame@gmail.com}

\and 
Huazhong Yang\\
Dept. of E.E.\\
Tsinghua University\\
{\tt\small yanghz@tsinghua.edu.cn}

\and
Yu Wang\footnote{*}\\
Dept. of E.E.\\
Tsinghua University\\
{\tt\small yu-wang@tsinghua.edu.cn}

}

\maketitle

{
  \renewcommand{\thefootnote}%
    {\fnsymbol{footnote}}
    \footnotetext[1]{Corresponding author.}
  \renewcommand{\thefootnote}%
    {\fnsymbol{footnote}}
    \footnotetext[1]{Preprint, work in progress.}
  }

\ificcvfinal\thispagestyle{empty}\fi

\begin{abstract}
   Adversarial attacks have rendered high security risks on modern deep learning systems. Adversarial training can significantly enhance the robustness of neural network models by suppressing the non-robust features. However, the models often suffer from significant accuracy loss on clean data. Ensemble training methods have emerged as promising solutions for defending against adversarial attacks by diversifying the vulnerabilities among the sub-models, simultaneously maintaining comparable accuracy as standard training. However, existing ensemble methods are with poor scalability, owing to the rapid complexity increase when including more sub-models in the ensemble. Moreover, in real-world applications, it is difficult to deploy an ensemble with multiple sub-models, owing to the tight hardware resource budget and latency requirement. In this work, we propose ensemble-in-one (EIO), a simple but efficient way to train an ensemble within one random gated network (RGN). EIO augments the original model by replacing the parameterized layers with multi-path random gated blocks (RGBs) to construct a RGN. By diversifying the vulnerability of the numerous paths within the RGN, better robustness can be achieved. It provides high scalability because the paths within an EIO network exponentially increase with the network depth. Our experiments demonstrate that EIO consistently outperforms previous ensemble training methods with even less computational overhead. 
\end{abstract}

\section{Introduction}
\label{pp:intro}

 With the convolutional neural networks (CNNs) becoming ubiquitous, the security and robustness of neural networks is attracting increasing focuses. Recent studies find CNN models are inherently vulnerable to adversarial attacks~\cite{goodfellow2014explaining}. These attacks can craft imperceptible perturbations 
 on the images, referred to as adversarial examples, to mislead the neural network models. Typical attack scenarios are often classified as 
 the white-box attack and 
 the black-box attack \cite{chakraborty2018adversarial}. A white-box attack occurs when an adversary can access the target model and has 
 full 
 knowledge of the weights, then they can generate 
 adversarial examples by fully exploring the most damaging perturbation noises 
 based on the known information. Otherwise, for a 
 black-box attack, the adversary cannot access the model. Alternatively, it 
 can generate adversarial examples from other surrogate models to attack the target model by exploiting the adversarial transferability among them. 

\begin{figure}
    \centering
    \includegraphics[scale=0.55]{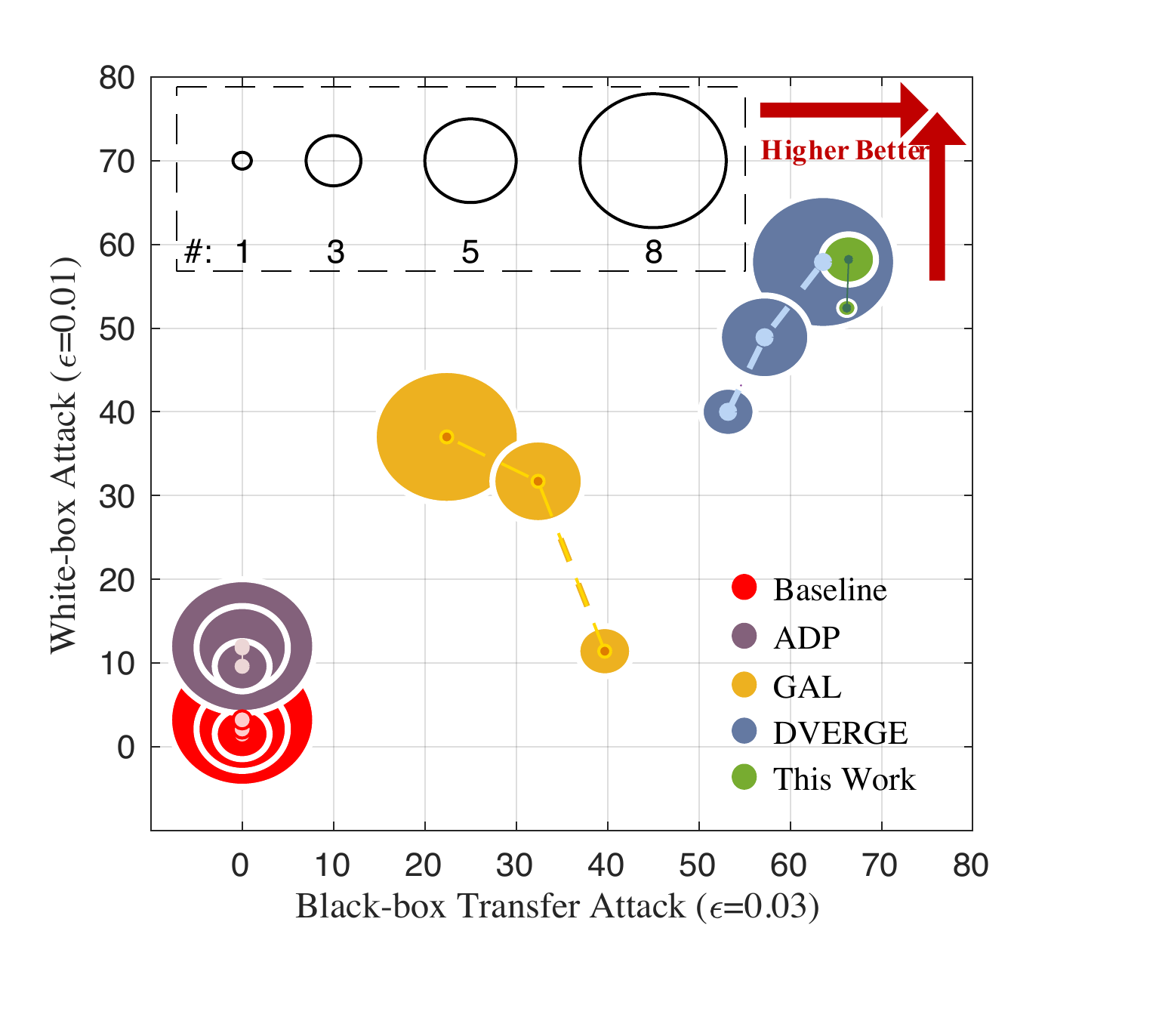}
    \vspace{-1.1cm}
    \caption{The overall accuracy comparison with state-of-the-art ensemble training methods. The $\#$ in the figure denotes the number of sub-models within the ensemble. Detail experimental setup can be found in Sec.\ref{pp:exp}. Our work consistently outperforms the previous methods without significant clean accuracy loss. Moreover, better robustness is achieved even with fewer sub-models within an ensemble, which greatly alleviates the computational pressure. }
    \label{fig:overall_perf}
  \end{figure}

Such vulnerability of CNN models has spurred extensive research on adversarial defenses. 
One stream of approach aims at learning robust features for an individual model \cite{madry2017towards, brendel2020adversarial}. 
Informally, robust features are defined as the features that are less sensitive to the perturbation noises added on the inputs. A representative approach, referred to as adversarial training \cite{madry2017towards}, on-line generates adversarial examples on which the model minimizes the training loss. 
As a result, adversarial training encourages the model to prefer robust features to non-robust features
, thereby alleviating the model's vulnerability. However, such adversarial training methods often significantly degrade the clean accuracy on the test dataset, since they exclude the non-robust features that usually have positive impacts on accuracy. 


Besides empowering improved robustness for an individual model, another stream of research focuses on designing methods to conduct strong \emph{ensembles} 
to defend against adversarial attacks \cite{yang2020dverge,bagnall2017training,pang2019improving,kariyappa2019improving}. The 
ensemble means the aggregation of multiple sub-models. Intuitively, an ensemble is expected to be more robust than an individual model because a successful attack needs to mislead the majority in the sub-models. The robustness of an ensemble highly relies on the diversity of vulnerabilities of the sub-models, then their decision boundaries will not intersect and be complementary.
 Motivated by this, many studies propose ensemble training methods to diversify the predictions 
 of the sub-models. For example, DVERGE \cite{yang2020dverge} distills the non-robust features corresponding to each sub-model's vulnerability. It isolates the vulnerability of the sub-models such that impeding 
 the transferability among them, thereby significantly improving the adversarial robustness without sacrificing the clean accuracy much. 

\begin{figure}
    \centering
    \includegraphics[scale=0.48]{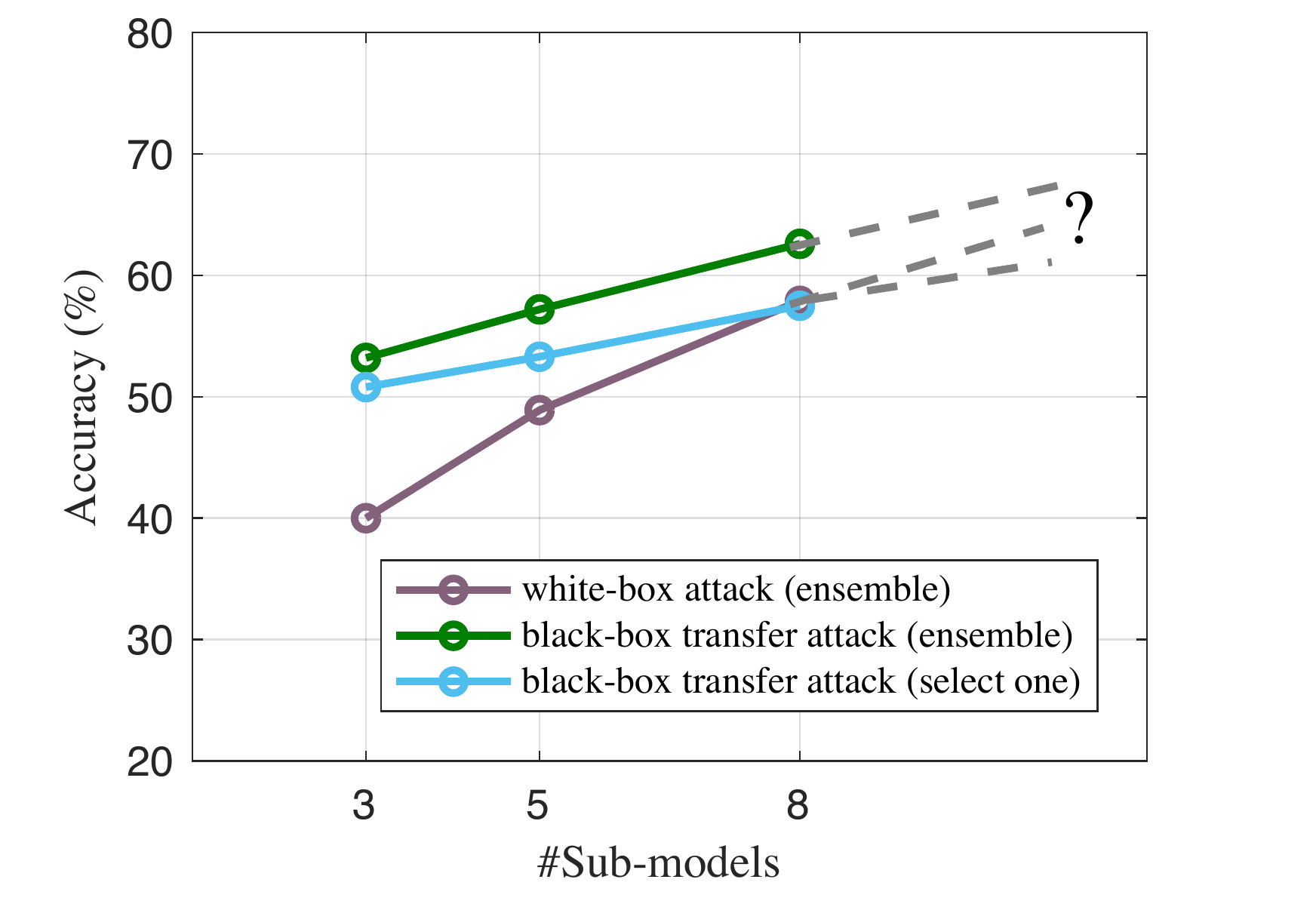}
    \caption{The trend of adversarial accuracies when the sub-models within an ensemble increase by leveraging DVERGE method \cite{yang2020dverge}. The perturbation strength for evaluating the black-box transfer attack and white-box attack is set to 0.03 and 0.01 respectively. Detailed experimental setup will be introduced in Sec.\ref{pp:exp}. The ``select one'' line represents the adversarial accuracy on an individual model selected from corresponding ensemble. }
    \label{fig:motivation_1}
\end{figure}


Despite recent work has has shown that ensembles composed by more sub-models tend to capture greater robustness improvement, these ensemble training methods are with poor scalability which hinders their broader applications.
Fig.\ref{fig:motivation_1} shows the robustness trend of the ensembles trained with the DVERGE method. Robustness improvement can be easily obtained by adding more sub-models into the ensemble. Meanwhile, when selecting an individual model from the 
ensembles respectively to test the accuracies under adversarial settings, similar trend can also be observed. 
However, it is hard to expand the scale of ensembles. 
We summarize the complexity of memory occupation, training and inference when scaling up $N$ in Table \ref{tab:scaleup}. 
For training, the complexity blow up significantly when $N$ enlarges. Especially in methods like DVERGE which train the sub-models in a round-robin manner, the training time will grow at the rate of $\mathcal{O}(N^2)$. Moreover, the memory requirement also become a hurdle for scaling up as it grows at the rate of $\mathcal{O}(N)$. Then the memory capacity of the training machine is probably insufficient to support simultaneous training of multiple sub-models, especially for large 
networks. For inference, it is practically infeasible to deploy an ensemble with multiple sub-models inside because they incur significant extra cost on the hardware resources and the running latency. 

\begin{table}[]
    \centering
    \begin{tabular}{|c|c|c|c|}
    \hline
       Method  & Memory & Training & Inference \\\hline\hline
       ADL/$N$ \cite{pang2019improving} & $\mathcal{O}(N)$ & $\mathcal{O}(N)$ & $\mathcal{O}(N)$ \\
       GAL/$N$ \cite{kariyappa2019improving} & $\mathcal{O}(N)$ & $\mathcal{O}(N)$ & $\mathcal{O}(N)$ \\
       DVERGE/$N$ \cite{yang2020dverge} & $\mathcal{O}(N)$ & $\mathcal{O}(N^2)$ & $\mathcal{O}(N)$ \\\hline
       Ours/$n^L$ & $<\mathcal{O}(n)$ & $\mathcal{O}(p^2)$ & $\mathcal{O}(1)$ \\\hline
       
    \end{tabular}
    \vspace{0.3cm}
    \caption{The complexity of memory, training, and inference w.r.t the number of sub-models $N$. The number after the slash in the first column stands for the instantiated sub-models. $n$ denotes the augmentation factor for each random gated block, $L$ denotes the depth of the networks, and $p$ denotes the samples of paths involved in each training iteration. Detailed explanation can be found in Sec.\ref{pp:method}.}
    \label{tab:scaleup}
\end{table}

Motivated by the aforementioned concerns, we propose \emph{Ensemble-in-One}, a novel approach that can improve the scalability of ensemble training, simultaneously obtaining better robustness and higher efficiency. For a dedicated model, we conduct a Random Gated Network (RGN) with auxiliary paths in each parameterized layer on top of the neural architecture. Through this, the network can instantiate numerous sub-models by randomly sample the paths. As concluded in Table \ref{tab:scaleup}, our method substantially reduce the complexity when scaling up the ensemble, as will explained in more detail in Sec.\ref{pp:exp}. We train the ensemble of paths within the one RGN and derive one individual path from the RGN for deployment, therefore we term the proposed method "Ensemble-in-One". In summary, the contributions of this work are listed as below:

\begin{itemize}
    \item Ensemble-in-One is a simple but effective method that learns adversarially robust ensembles within one over-parametrized random gated network. The EIO construction enables us to employ ensemble learning techniques to learn more robust individual models with minimal computational overheads  and no extra inference overhead. 
    \item Extensive experiments demonstrate the effectiveness of Ensemble-in-One. It consistently outperforms the previous ensemble training methods with negligible accuracy loss. As shown in Fig.\ref{fig:overall_perf}, Ensemble-in-One achieves even better robustness than 8-sub-model ensembles trained by previous methods with only one individual model. 
\end{itemize}

\section{Related Work}
\label{pp:relate_work}

\subsection{Adversarial attacks and countermeasures.} The inherent vulnerability of CNN models poses challenges on the security of deep learning systems. An adversary can apply an additive perturbation on an original input, which is usually imperceptible to human, to generate an adversarial example that induces wrong prediction in CNN models \cite{goodfellow2014explaining}. Denoting an original input as $x$, the goal of adversarial attacks is to find a perturbation $\delta$ s.t. $x_{adv}=x+\delta$ can mislead the model and $||\delta||_p$ satisfies the intensity constraint $||\delta||_p \leq \epsilon$. To formulate that, the adversarial attack aims at maximizing the loss $\mathcal{L}$ for the model with parameters $\theta$ on the input-label pair $(x,y)$, i.e. $\delta=\mathrm{argmax}_{\delta} \mathcal{L}_{\theta}(x+\delta,y)$, under the constraint that the $\ell_p$ norm of the perturbation should not exceed the bound $\epsilon$: $||\delta||_p \leq \epsilon$.  Usually, we use $\ell_\infty$ norm \cite{goodfellow2014explaining, madry2017towards} of the perturbation intensity to measure the attack strength or model's robustness. An attack that requires smaller perturbation to successfully deceive the model is regarded to be stronger. Correspondingly, a defense that forces the attack to enlarge perturbation intensity is regarded to be more robust.

Various adversarial attack methods have been investigated to strengthen the attack effectiveness. The fast gradient sign method (FGSM) \cite{goodfellow2014explaining} utilizes the gradient descent method to generate adversarial examples. As an improvement, many studies further show the attack can be strengthened through multi-step projected gradient descent (PGD) \cite{madry2017towards} generation, random-starting strategy, and momentum mechanism \cite{dong2017discovering}. Then SGM \cite{wu2020skip} further finds that adding weight to the gradient through the skip connections can make the attacks more effective. Other prevalent attack approaches include C\&W \cite{carlini2017towards}, M-DI$^2$-FGSM \cite{xie2019improving}, etc. These attacks provide strong and effective ways to generate adversarial examples, rendering a huge threat to real-world deep learning systems. 

To improve the robustness of CNN systems, there are also extensive countermeasures for adversarial attacks. One active research direction targets improving the robustness of individual models. Adversarial training \cite{madry2017towards} optimizes the model on the adversarial examples generated in every step of the training stage. Therefore, the optimized model will tend to drop non-robust features to converge better on the adversarial data. However, adversarial training encourages the model to fit the adversarial examples, thereby reducing the generalization on the clean data and causing significant degradation of the clean accuracy. 

\subsection{Test-time randomness for adversarial defense}
Besides the aforementioned training techniques, there exist studies that introduce test-time randomness to improve the model robustness. Feinman et. al.~\cite{feinman2017detecting} utilize the uncertainty measure in dropout networks to detect adversarial examples. Dhillon et. al.~\cite{Dhillon2018stochastic} and Xie et. al.~\cite{xie2017mitigating} incorporate layer-wise weighted dropout and random input transformations during test time to improve the robustness.
Test-time randomness is found to be effective in increasing the required distortion on the model, since test-time randomness makes generating white-box adversarial examples almost as difficult as generating transferable black-box ones~\cite{Carlini2017adversarial}. Nevertheless, test-time randomness increases the inference cost and can be circumvented to some extent with the expectation-over-transformation technique~\cite{athalye2018obfuscated}. 

\subsection{Ensemble training for adversarial defense.} 
Besides improving the robustness of individual models, another recent research direction is to investigate the robustness of model ensembles in which multiple sub-models work together. The basic idea is that multiple sub-models can provide diverse decisions. Similar to bagging \cite{breiman1996bagging} and boosting \cite{dietterich2000ensemble}, ensemble methods can combine multiple weak models to jointly make decisions, thereby assembling as a stronger entirety. However, independent training leads to similar feature representations, which would not provide diversities among the sub-models \cite{kariyappa2019improving}. Therefore, several studies propose ensemble training methods to fully diversify the features representation to impede the transferability among the sub-models and improve the ensemble robustness. Pan et. al. propose an adaptive diversity promoting (ADP) regularizer \cite{pang2019improving} to encourage the diversity among the individual models. Sanjay et. al. propose a gradient alignment loss (GAL) \cite{kariyappa2019improving} which takes the cosine similarity of the gradients to approximate the coherence of sub-models. The very recent work DVERGE exploits feature distillation to diversify the vulnerabilities among the sub-models. By learning from the non-robust features distilled from the sub-models, DVERGE \cite{yang2020dverge} successfully isolate and diversify the vulnerability in each sub-model such that the within-ensemble transferability is highly impeded. Thus, DVERGE achieves improved robustness without significantly impacting the clean accuracy.

\begin{figure}
    \centering
    \includegraphics[scale=0.48]{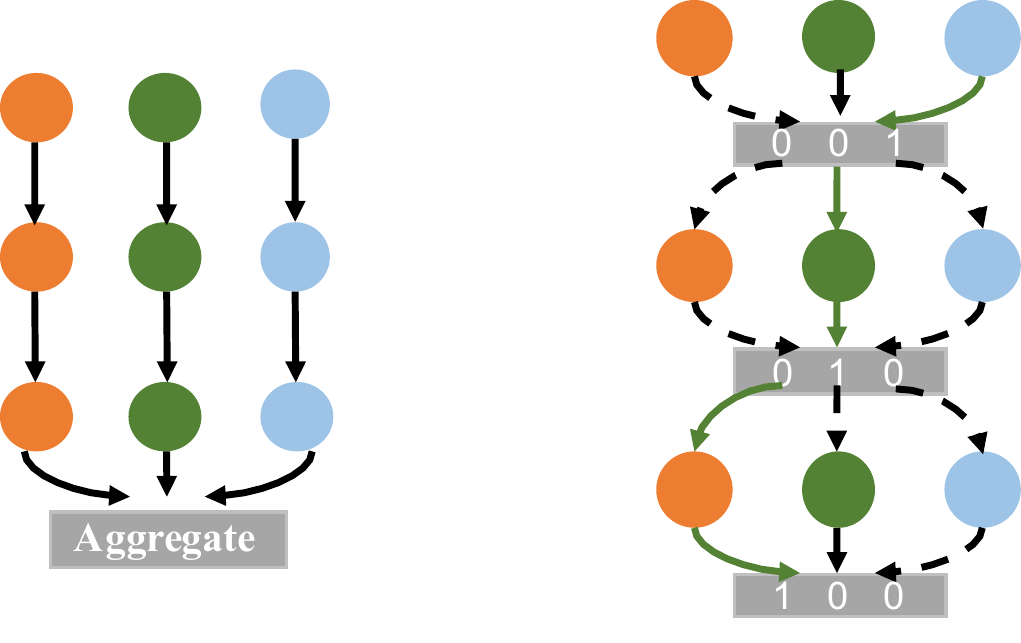}
    \caption{Normal ensemble training of multiple sub-models (left) and the proposed ensemble-in-one training within a random gated network (right). By selecting the paths along augmented layers, the ensemble-in-one network can instantiate $n^L$ sub-models, where $n$ represents the augmentation factor of the multi-gated block for each augmented layer and $L$ represents the number of augmented layers in the network.}
    \label{fig:ensemble_in_one}
\end{figure}

\begin{figure*}
    \vspace{-0.4cm}
    \centering
    \includegraphics[scale=0.44]{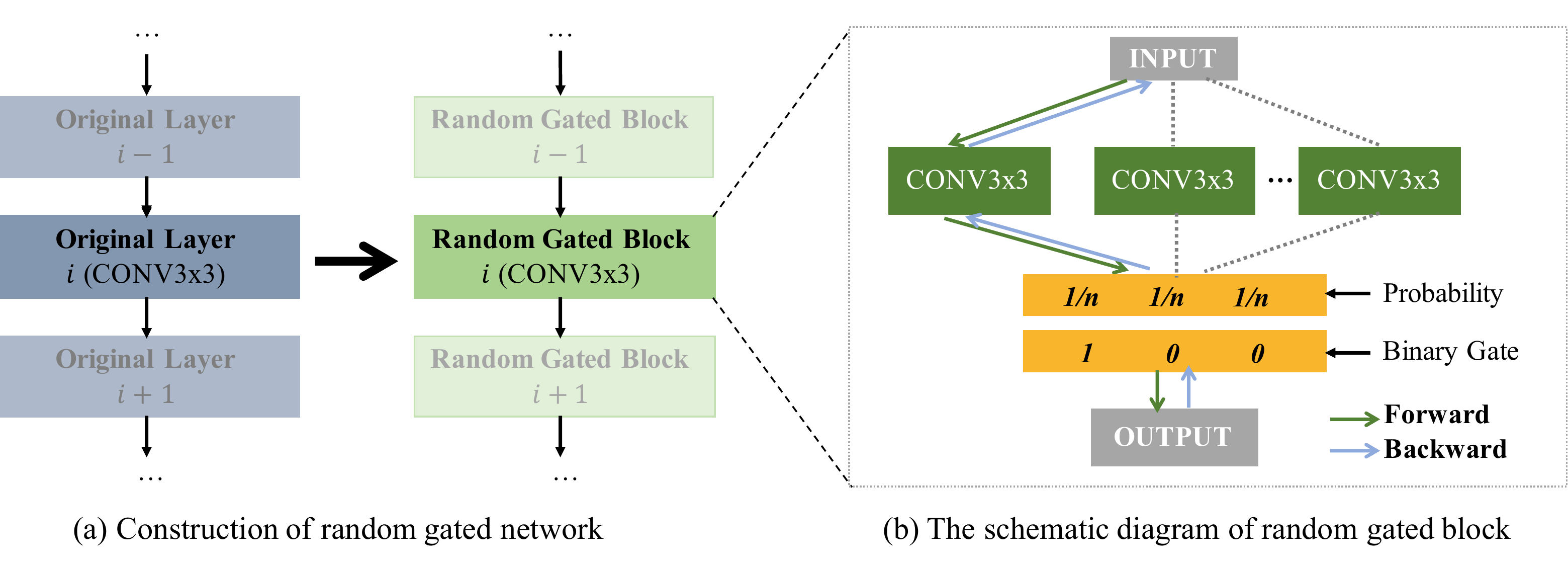}
    \caption{The construction of random gated network based on random gated blocks. The forward propagation will select one path to allow the input pass. Correspondingly, the gradients will also propagate backward along the same path.} 
    \label{fig:dynamic_block}
\end{figure*}


\section{Ensemble-in-One}
\label{pp:method}

In this section, we first introduce the basic motivation of our approach. Then we introduce the construction of the random gated network (RGN) with basic random gated blocks (RGBs). Then we propose a training algorithm to learn an ensemble within the RGN by leveraging existing diversity optimization methods. Finally, we further discuss the derivation and deployment strategies from the RGN. 

\subsection{Basic Motivation}
As illustrated in Sec.\ref{pp:intro}, the conventional way to augment ensembles is to aggregate multiple sub-models, which is inefficient and hard to scale up. An intuitive way to enhance the scalability of the ensemble construction is to introduce an ensemble for each later in the network. As shown in Fig.\ref{fig:ensemble_in_one}, we can augment a dynamic network by augmenting each parameterized layer with an $n$-path gated block. Then by selecting the paths along the augmented layer, the dynamic network can instantiate $n^L$ varied sub-models ideally. These paths are expected to provide numerous vulnerability diversities. Taking ResNet-20 as an example, by replacing each convolutional layer with a two-path gated module, the overall paths will approach $2^{21}$. Such augmentation provides an approximation to training a very large ensemble of sub-models. Then through vulnerability diversification cross-training, each path tends to capture better robustness. Following this idea, we propose \emph{Ensemble-in-One} to further improve the robustness of both individual models and ensemble models. 

\subsection{Construction of the Random Gated Network}
Denote a candidate neural network as $\mathcal{N}(o_1, o_2, ..., o_m)$, where $o_i$ represents an operator in the network. To transform the original network into a random gated network, we first extract the neural architecture to obtain the connection topology and operation types. On top of that, 
we replace each parameterized layer (mainly convolutional layer, optionally followed by a batch normalization layer) with a random gated block (RGB). As shown in Fig.~\ref{fig:dynamic_block}, each RGB simply repeats the original layer by $n$ times, and leverages binary gates with the same probabilities to control the open or shutdown of corresponding sub-layers. These repeated sub-layers share different parameters.
We denote the random gated network (RGN) as $\mathcal{N}(d_1, d_2, ..., d_m)$, where $d_i=(o_{i1}, ..., o_{in})$. Let $g_i$ be the gate information in the $i_{\rm{th}}$ RGB, then a specific path derived from the RGN can be expressed as $\mathcal{P}=(g_1\cdot d_1, g_2\cdot d_2, ..., g_m\cdot d_m)$. 

For each RGB, when performing the computation, only one of the $n$ gates is opened at a time, and the others will be temporarily pruned. Thus by, only one path of activation is active in memory during training, which reduces the memory occupation of training an RGN to the same level of training an individual model. 
Moreover, to ensure that all paths can be equally sampled and trained, each gate in a RGB is chosen with identical probability, i.e. $1/n$ if each RGB consists of $n$ sub-operators. Therefore, the binary gate function can be expressed as:
\vspace{-0.2cm}
\begin{equation}
    \begin{aligned}
      g_i =
      \begin{cases}
      [1, 0, ..., 0] \quad \text{with probability $1/n$}, \\
      [0, 1, ..., 0] \quad \text{with probability $1/n$}, \\
      \quad \quad \text{...} \\
      [0, 0, ..., 1] \quad \text{with probability $1/n$}. \\
      \end{cases}
    \end{aligned}
    \label{eq:gate}
\end{equation}
An RGN is analogous to the super network in parameter-sharing neural architecture search, and the forward process of an RGN is similar to evaluating a sub-architecture~\cite{pham2018efficient,cai2018proxylessnas}. Compared to conventional ensemble training methods, our method is easier to scale up the ensemble. It only incurs $n\times$ memory occupation for the weight storage, while still keeping the same memory requirement for activation as an individual model. 

\subsection{Learning Ensemble in One}
The goal of learning ensemble-in-one is to encourage the vulnerabilities diversity of all the paths within the RGN by round-robinly learning from each other. Let $\mathcal{P}_i$ and $\mathcal{P}_j$ be two different paths, 
 where we define two paths as different when at least one of their gates is different. To diversify the vulnerabilities, we need first distill the non-robust features of the paths so that the optimization process can isolate them. We adopt the same feature distillation objective as previous work \cite{ilyas2019adversarial,yang2020dverge}. 
 Consider two independent input-label pairs $(x_t,y_t)$ and $(x_s,y_s)$ from the training dataset, the distilled feature of $x_t$ corresponding to $x_s$ by the $l_{\rm{th}}$ layer of path $\mathcal{P}_i$ can be achieved by:
\begin{equation}
    x'_{\mathcal{P}_i^l}(x_t, x_s) = \text{argmin}_z||f_{\mathcal{P}_i}^l(z) - f_{\mathcal{P}_i}^l(x_t)||^2,
    \label{eq:distill}
\end{equation}
where $||z-x_s||_{\infty} \leq \epsilon_d$. Such feature distillation aims to construct a sample $x'_{\mathcal{P}_i^l}$ by adding slight perturbation on $x_s$ so that the feature response of $l_{\rm{th}}$ layer of $\mathcal{P}_i$ on $x'_{\mathcal{P}_i^l}$ is similar as $x_t$, while the two inputs $x_t$ and $x_s$ are completely independent. This exposes the vulnerability of path $\mathcal{P}_i$ on classifying $x_s$. Therefore, for another different path $\mathcal{P}_j$, it can learn on the distilled data to correctly classify them to circumvent the vulnerability. The optimization objective for path $\mathcal{P}_j$ is to minimize:
\begin{equation}
    \mathbb{E}_{(x_t, y_t), (x_s, y_s),l}\mathcal{L}_{f_{\mathcal{P}_j}}(x'_{\mathcal{P}_i^l}(x_t, x_s), y_s).
\end{equation}
As it is desired that each path can learn from the vulnerabilities of all the other paths, the objective of training the ensemble-in-one RGN is to minimize:
\begin{equation}
    \sum_{\forall \mathcal{P}_j \in \mathcal{N}}\mathbb{E}_{(x_t, y_t), (x_s, y_s),l}\sum_{\forall \mathcal{P}_i \in \mathcal{N}, i\neq j}\mathcal{L}_{f_{\mathcal{P}_j}}(x'_{\mathcal{P}_i^l}(x_t, x_s), y_s),
\end{equation}
where $\mathcal{N}$ is the set of all paths in the RGN. While it is obviously impossible to involve all the paths in a training iteration, we randomly sample a certain number of paths by stochastically set the binary gates according to Eq.\ref{eq:gate}. We denote the number of paths sampled in each iteration as $p$. Then the selected paths can temporarily combine as a subset of the RGN, referred to as $\mathcal{S}$. The paths in the set $\mathcal{S}$ keep changing throughout the whole training process, such that all paths will have equal opportunities to be trained.

The training process of the RGN is summarized by the pseudo-code in Algorithm \ref{alg:routine}. Before starting vulnerability diversification training, we pre-train the RGN based on standard training settings to help the RGN obtain basic capabilities. The process is simple, where a random path will be sampled in each iteration and trained on clean data. Then for each batched data, the process of vulnerability diversification contains three basic steps. First, random sampling of $p$ paths to be involved in the iteration. Note that the sampled paths should be varied, i.e. if the distilling layer is set to $l$, for any $\mathcal{P}_i$, $\mathcal{P}_j$ in $\mathcal{S}$, there must be at least one different gate among the top $l$ gates, i.e. $\exists k \in [1, l]$, s.t. $\mathcal{P}_i[k] \neq \mathcal{P}_j[k]$. Second, distilling the vulnerable features of the sampled paths according to Eq. \ref{eq:distill}. The distillation process is the same as proposed in DVERGE, by applying a PGD scheme for approximating the optimal adversarial data. Third, train each path with the distilled data from the other paths in a round-robin manner. Because the paths unavoidably share a proportion of weights owing to the weight sharing mechanism, the gradients of the weights will not be updated until all sampled paths are included. 


\subsection{Model Derivation and Deployment}
Once the training of RGN is finished, we can then derive and deploy the model in two ways. One way is to deploy the entire RGN, then in inference stage, the gates throughout the network will be randomly selected to process an input. The advantage is that the computation is randomized, which may beneficial for improving the robustness under white-box attacks, because the transferability among different paths was impeded during diversity training. However, the disadvantage is that the accuracy is unstable owing to the dynamic choice of inference path, where the fluctuation reaches 1-2 percentage. 

Another way is to derive individual models from the RGN. By sampling a random path and eliminating the other redundant modules, an individual model can be rolled out. We can also sample multiple paths and derive multiple models to combine as an ensemble. Deploying models in this way ensures the stability of the prediction as the randomness is eliminated. In addition, the derived models can be slightly finetuned with small learning rate for a few epochs to compensate for the under-convergence, as the training process of RGN cannot fully train all paths as the probability of each specific path being sampled is relatively low.

\begin{figure}[!tt]
\vspace{-0.2cm}
\begin{algorithm}[H] \footnotesize 
    \caption{{\small Training process for learning Ensemble-in-One}}
	\label{alg:routine}
	\begin{algorithmic}[1]
	    \Require Path samples per ietration $p$
	    \Require Random Gated Network $\mathcal{N}$ with $L$ parameterized layers 
	    \Require Pre-training epoch $E_w$, training epoch $E$, and data batch $B_d$
	    \Require Optimization loss $\mathcal{L}$, learning rate $lr$
	    \Ensure Trained Ensemble-in-One model \\
	    
	    \text{\# pre-training of $\mathcal{N}$}
	    \For{e = 1, 2, ..., $E_w$}
	    \For{b = 1, 2, ..., $B_d$}
	    \State \text{Random Sample Path $\mathcal{P}_i$ from $\mathcal{N}$}
	    \State \text{Train $\mathcal{P}_i$ in batched data}
	    \EndFor
	    \EndFor \\
	    \text{\# learning vulnerability diversity for $\mathcal{N}$}
	    \For{e = 1, 2, ..., $E$)}
	    \For{b = 1, 2, ..., $B_d$)}
	    \State Random sample $l\in [1, L]$ 
	    \State \text{\# randomly sample $p$ paths}
	    \State $\mathcal{S}$=[$\mathcal{P}_1$, $\mathcal{P}_2$, ..., $\mathcal{P}_{p}$], s.t. $\forall i, j, \exists k \in [1, l]$, s.t. $\mathcal{P}_i[k] \neq \mathcal{P}_j[k]$
	    \State Get data $(X_t, Y_t), (X_s, Y_s)$ $\leftarrow$ $D$
	    \State \# Get distilled data

	    \For{i = 1, 2, ..., $p$}
	    \State $X_i' = x'_{\mathcal{P}_i^l}(X_t, X_s)$ 
	    \EndFor
	    \State $\nabla_{\mathcal{N}} \leftarrow 0$
	    \For{i = 1, 2, ..., $p$}
	    \State $ \nabla_{\mathcal{P}_i} = \nabla( \sum_{j\neq i}\mathcal{L}_{f_{\mathcal{P}_i}}(f_{\mathcal{P}_i}(X_j'), Y_s))$
	    \State $\nabla_{\mathcal{N}} = \nabla_{\mathcal{N}} + \nabla_{\mathcal{P}_i}$
	    \EndFor
	    \State $\mathcal{N} = \mathcal{N} - lr * \nabla_{\mathcal{N}}$
	    \EndFor
	    \EndFor
	   

	\end{algorithmic}
\end{algorithm}
\vspace{-0.5cm}
\end{figure}


\section{Experimental Results}
\label{pp:exp}

\subsection{Experiment Settings}
\textbf{Benchmark.} The experiments are constructed on the ResNet-20 network \cite{he2016deep} with the CIFAR-10 dataset \cite{krizhevsky2009learning}. Specifically, we construct the ResNet-20-based RGN by transforming each convolution layer to a two-path RGB (in default). Overall, there are 21 RGBs (containing 19 convolution layers in the straight-through branch and two convolution layers in the skip connection branch). To evaluate the effectiveness of our method, we compare Ensemble-in-One with four counterparts, including the \emph{Baseline} which trains the models in a standard way and three previous ensemble training methods: \emph{ADL} \cite{pang2019improving}, \emph{GAL} \cite{kariyappa2019improving}, and \emph{DVERGE} \cite{yang2020dverge}. 

\textbf{Training Details.} The trained ensemble models of baseline, ADL, GAL, and DVERGE are downloaded from the public repository released in \cite{yang2020dverge}. We train the Ensemble-in-One network for 200 epochs using SGD with momentum 0.9 and weight decay 0.0001. The initial learning rate is 0.1, and decayed by 10x at the 100-th and the 150-th epochs respectively. When deriving the individual models, we fine-tune the derived models for 40 epochs using SGD with momentum 0.9 and weight decay 0.0001. The initial learning rate is 0.001, and decayed by 10x at the 20-th and 30-th epochs respectively. In default, for the RGN training, we sample 3 paths per iteration. The augmented factor for each RGB is set to 2, and the PGD-based perturbation strength $\epsilon_d$ for feature distillation is set to 0.07 with 10 iterative steps and each step size of $\epsilon_d/10$. 

\begin{figure}
    \centering
    \includegraphics[scale=0.45]{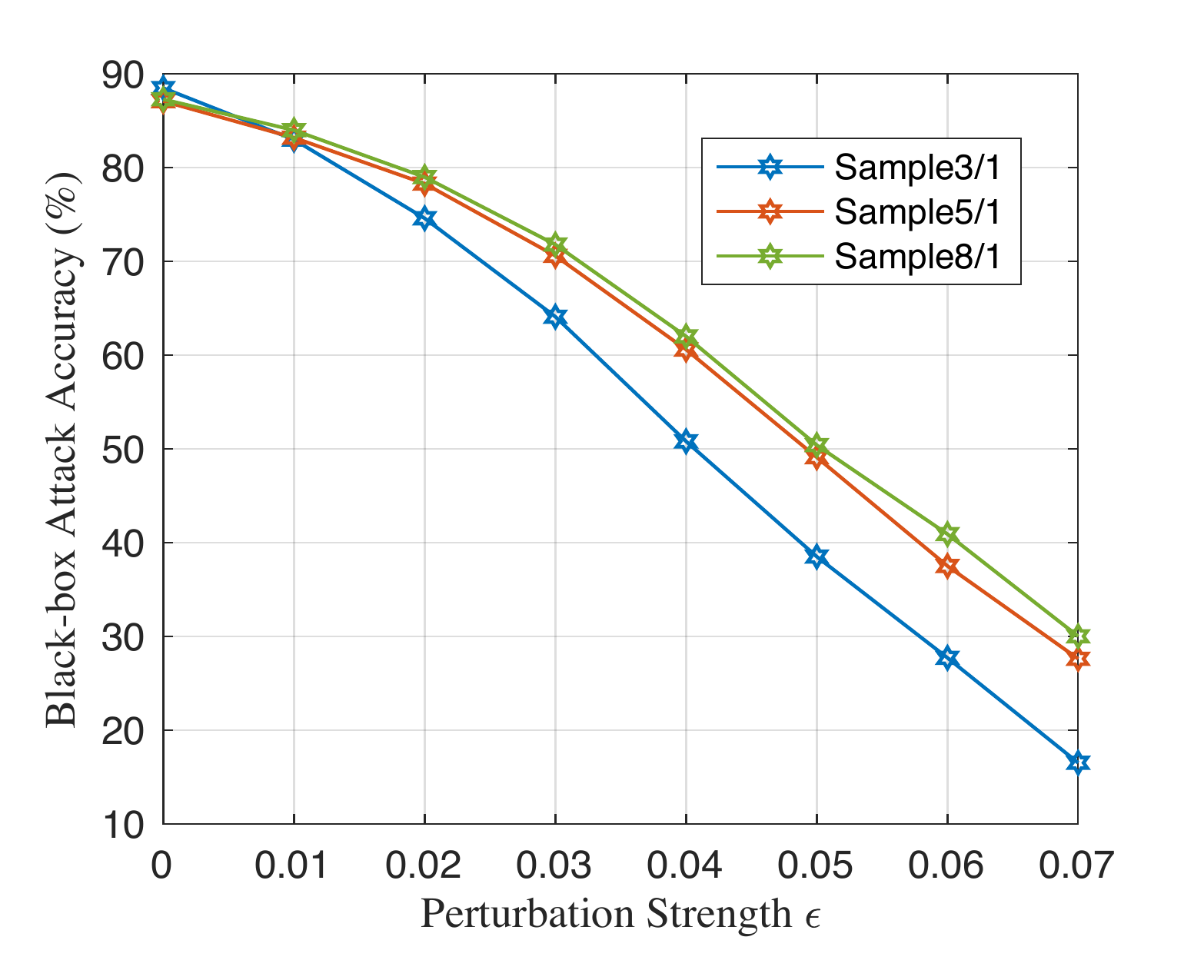}
    \vspace{-0.2cm}
    \caption{The adversarial accuracy versus perturbation strength under black-box transfer attacks with different path batchsize as mentioned in Algorithm \ref{alg:routine}. The number after the slash stands for the number of models derived from the RGN. And the number after ``Sample'' stands for the path samples in each training iteration.  }
    \label{fig:batch}
\end{figure}

\begin{figure}
    \centering
    \includegraphics[scale=0.47]{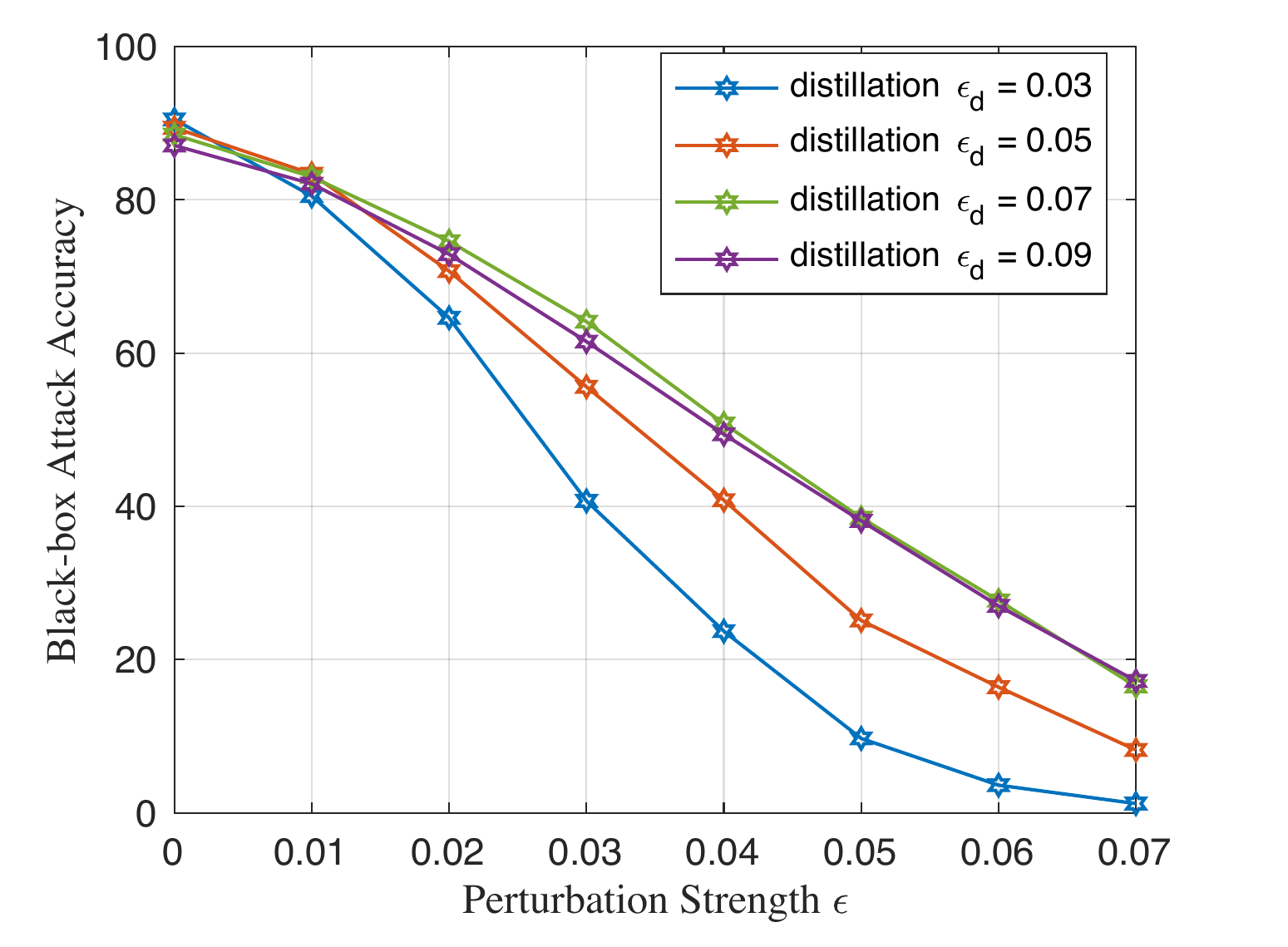}
    \vspace{-0.2cm}
    \caption{The adversarial accuracy versus perturbation strength under black-box transfer attacks with different distillation $\epsilon_d$ as mentioned in Eq.\ref{eq:distill}. The curves covers a wide range of distillation $\epsilon_d$ from 0.03 to 0.09. }
    \label{fig:eps}
\end{figure}

\begin{figure*}
    \hspace{-0.2cm}
    \vspace{-0.2cm}
    \includegraphics[scale=0.6]{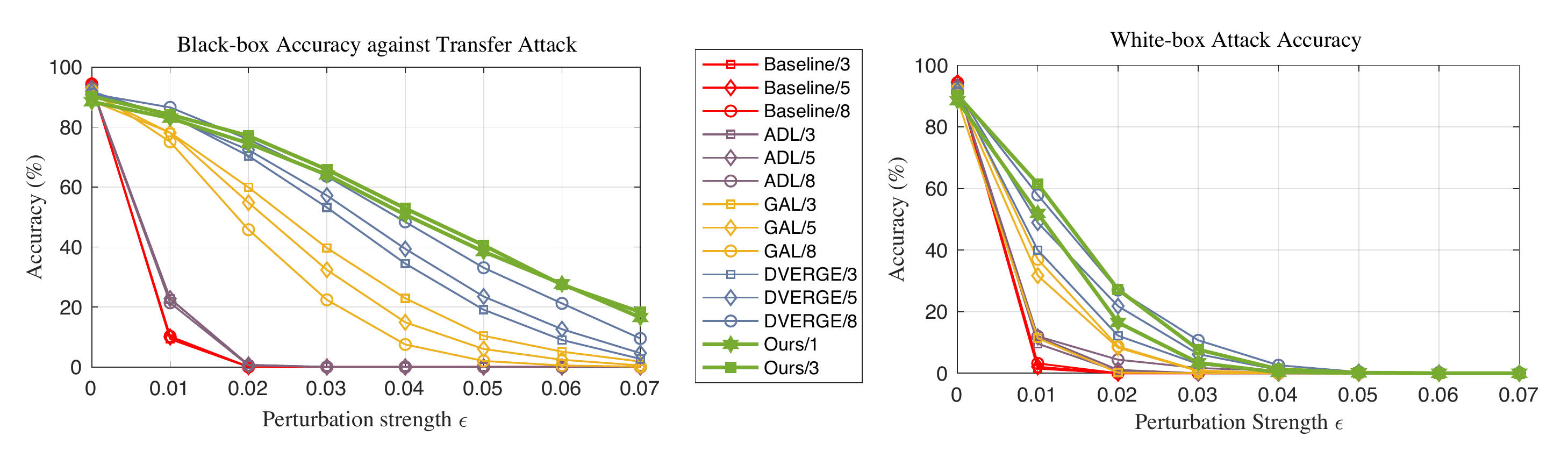}
    
    \caption{Contrasting the robustness of Ensemble-in-One with previous ensemble training methods. Left: adversarial accuracy under black-box transfer attack; and right: adversarial accuracy under white-box attack. The number after the slash stands for the number of sub-models within the ensemble. }
    \label{fig:perf_compare}
\end{figure*}{}

\textbf{Attack Models.} We categorize the adversarial attacks as black-box transfer attacks and white-box attacks. As illustrated in Sec.\ref{pp:intro}, the white-box attack assumes the adversary has full knowledge of the target model parameters and architectures, and the black-box attack assumes the adversary cannot access the parameters and can only generate adversarial examples from surrogate models to transfer attack the target model. For fair comparison, we adopt exactly the same attack methodologies and the same surrogate models as DVERGE to evaluate the robustness. For black-box transfer attacks, the attack methods include: (1) PGD with momentum and with three random starts \cite{madry2017towards}; (2) M-DI$^2$-FGSM \cite{xie2019improving}; and (3) SGM \cite{wu2020skip}. The attacks are with different perturbation strength and the iterative steps are set to 100 with the step size of $\epsilon$/5. Besides the cross-entropy loss, we also apply the C\&W loss to incorporate with the attacks. Therefore, there will be 3 (surrogate models) $\times$ 5 (attack methods, PGD with three random starts, M-DI$^2$-FGSM, and SGM) $\times$ 2 (losses) = 30 adversarial attacks. For white-box attacks, we apply 50-step PGD with the step size of $\epsilon/5$ with five random starts. Both the black-box and white-box adversarial accuracy is reported in a \emph{all-or-nothing} fashion: a sample is judged to be correctly classified only when its 30 (for black-box transfer attack) or 5 (for white-box attack) adversarial versions are all corrected classified by the model. In default, we randomly sample 1000 instances from the CIFAR-10 test dataset to evaluate the accuracy. We believe the attacks are powerful and can distinguish the robustness of the various models.

\subsection{Robustness Evaluation}
\textbf{Hyper-parameter Exploration.} Recall that three important hyper-parameters are involved in the training procedure. One is the number of sampled paths $p$ to participate in each training iteration, one is the strength of feature distillation perturbation $\epsilon_d$ as illustrated in Eq.\ref{eq:distill}, and the other is the augmentation factor $n$ for constructing the RGN, i.e. how many times will an operator be repeated to build a RGB. We make experiments to empirically explore the optimal hyper-parameters for better trading-off the clean accuracy and the adversarial accuracy. 

Fig.\ref{fig:batch} shows the curves of black-box adversarial accuracy under different sampled path number $p$. As is observed, when the sampled paths increase, the robustness of the derived individual model also improves. The underlying reason is that more samples of paths participating in each iteration allows more paths to be cross-trained, thereby each path is expected to learn from more diverse vulnerabilities. However, the clean accuracy slightly drops with the increasing of path samples, and the training time will increase as the complexity is $\mathcal{O}(p^2)$. Hence, sampling 3 paths per iteration will be a relatively optimal choice.

Fig.\ref{fig:eps} shows the curves of black-box adversarial accuracy under different feature distillation $\epsilon_d$. We find similar conclusions as presented in DVERGE. A larger $\epsilon_d$ can push the distilled data $x'_{\mathcal{P}_i^l}(x_t, x_s)$ share more similar internal representation as $x_t$. While the objective is to reduce the loss of $\mathcal{P}_j$ on classifying $x'_{\mathcal{P}_i^l}$, the larger loss will boost the effectiveness of learning the diversity, thereby achieving better robustness. However, we also find the clean accuracy drops with the increase of $\epsilon_d$. And there exists a switching point where it will stop obtaining robustness improvement from continually increasing $\epsilon_d$. The experimental results suggest $\epsilon_d=0.07$ to achieve higher robustness and clean accuracy simultaneously. 

\begin{table}[]
    \centering
    \begin{tabular}{c|c|ccc}
    \hline
      \#Sub-model & $n$ & Clean & Black-box & White-box \\\hline\hline
       1  & 2 & 88.5\% & 64.1\% & 51.9\%\\
       1  & 3 & 88.8\% & 61.6\% & 48.2\% \\\hline
       3  & 2 & 90.3\% & 65.9\% & 61.5\% \\
       3  & 3 & 89.1\% & 62.9\% & 53.3\% \\ \hline
    \end{tabular}
    \vspace{0.2cm}
    \caption{The comparison of different augmentation factor $n$ for the RGN. The adversarial accuracy under black-box attack and white-box attack are evaluated with $\epsilon=0.03$ and $\epsilon=0.01$ respectively. }
    \label{tab:n}
\end{table}

Table \ref{tab:n} shows the comparison of adversarial accuracy when applying different augmentation factor $n$ for constructing the RGN. Observe that increasing the factor $n$ brings no benefit on either the clean accuracy or adversarial accuracy. It stands to reason that augmenting $2\times$ operators for each RGB has already provided sufficient random paths. Moreover, increasing the $n$ may lead to more severe under-convergence of training because each path has a decreased probability of being sampled. To conclude that, we set the hyper-parameters as $\epsilon_d$=$0.07$, $p$=$3$, $n$=$2$. We keep these hyper-parameter settings in following experiments.

\textbf{Comparison with Other Ensemble Methods.} Fig.\ref{fig:perf_compare} shows the overall adversarial accuracy of the models trained by different methods with a wide range of attack perturbation strength. The results show that through our Ensemble-in-One method, an individual model derived from the RGN can significantly outperform the heavy ensembles trained by previous methods with higher adversarial accuracy under both black-box and white-box attacks, simultaneously achieving comparable clean accuracy. The results demonstrate that we successfully realize the ensemble-in-one vision as illustrated in Sec.\ref{pp:intro}, i.e. training an ensemble within one network and improves the robustness of an individual model to outperform the ensembles such that the deployment overhead can be substantially reduced.

\begin{figure}
    \hspace{-0.3cm}
    \vspace{-0.1cm}
    \includegraphics[scale=0.4]{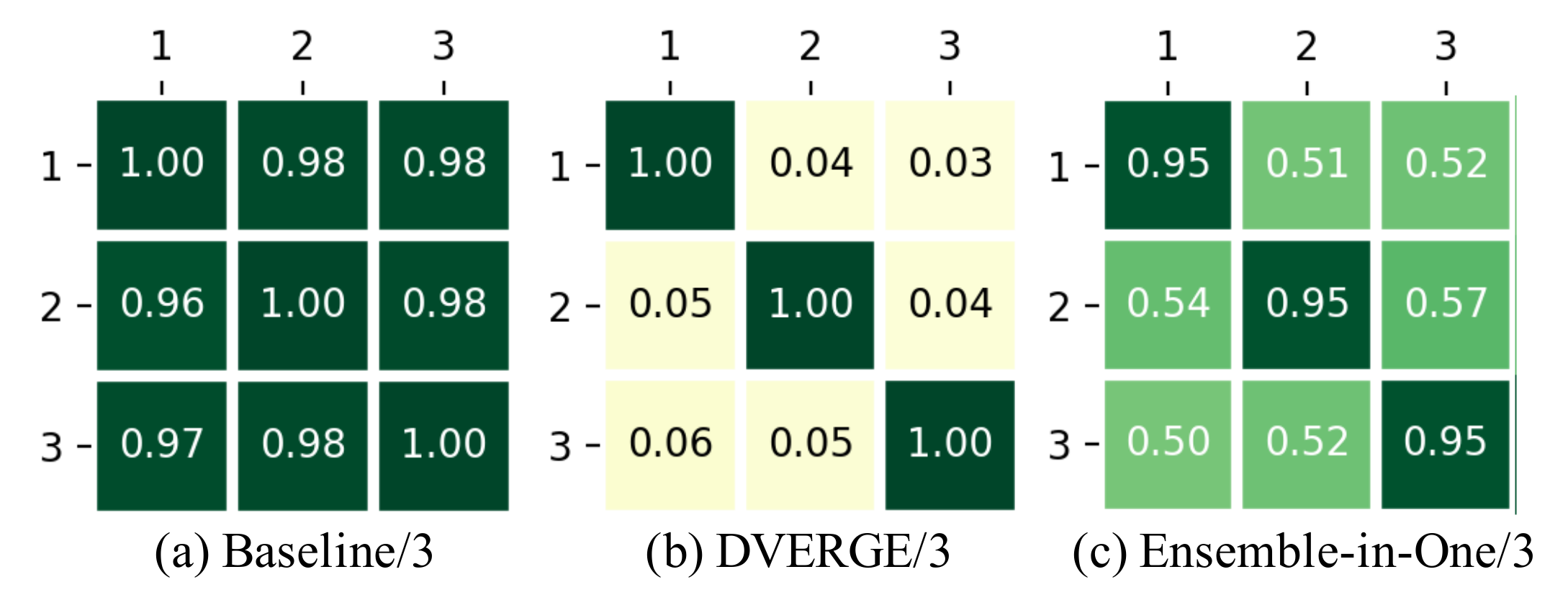}
    \caption{The transferability among the sub-models within corresponding ensemble evaluated with $\epsilon=0.03$. The transferability is evaluated in the form of attack success rate. The number after the slash represents the number of sub-models within the ensemble.}
    \label{fig:transfer}
\end{figure}

\textbf{Transferability Evaluation.} Fig.\ref{fig:perf_compare} also points out that the trend toward improving robustness by increasing sub-models within the ensemble is not as obvious as observed in the DVERGE method. The underlying reason is that the transferability among different paths within the RGN is not completely impeded, owing to the weight sharing mechanism of RGN training. As shown in Fig.\ref{fig:transfer}, although Ensemble-in-One captures lower transferability among the sub-models than the Baseline method, it is still far higher than DVERGE. This also leads to poor complementarity among the paths, which makes it hard to obtain better robustness by combining multiple paths as an ensemble. 

\begin{figure}
    \hspace{-0.3cm}
    \vspace{-0.1cm}
    \includegraphics[scale=0.42]{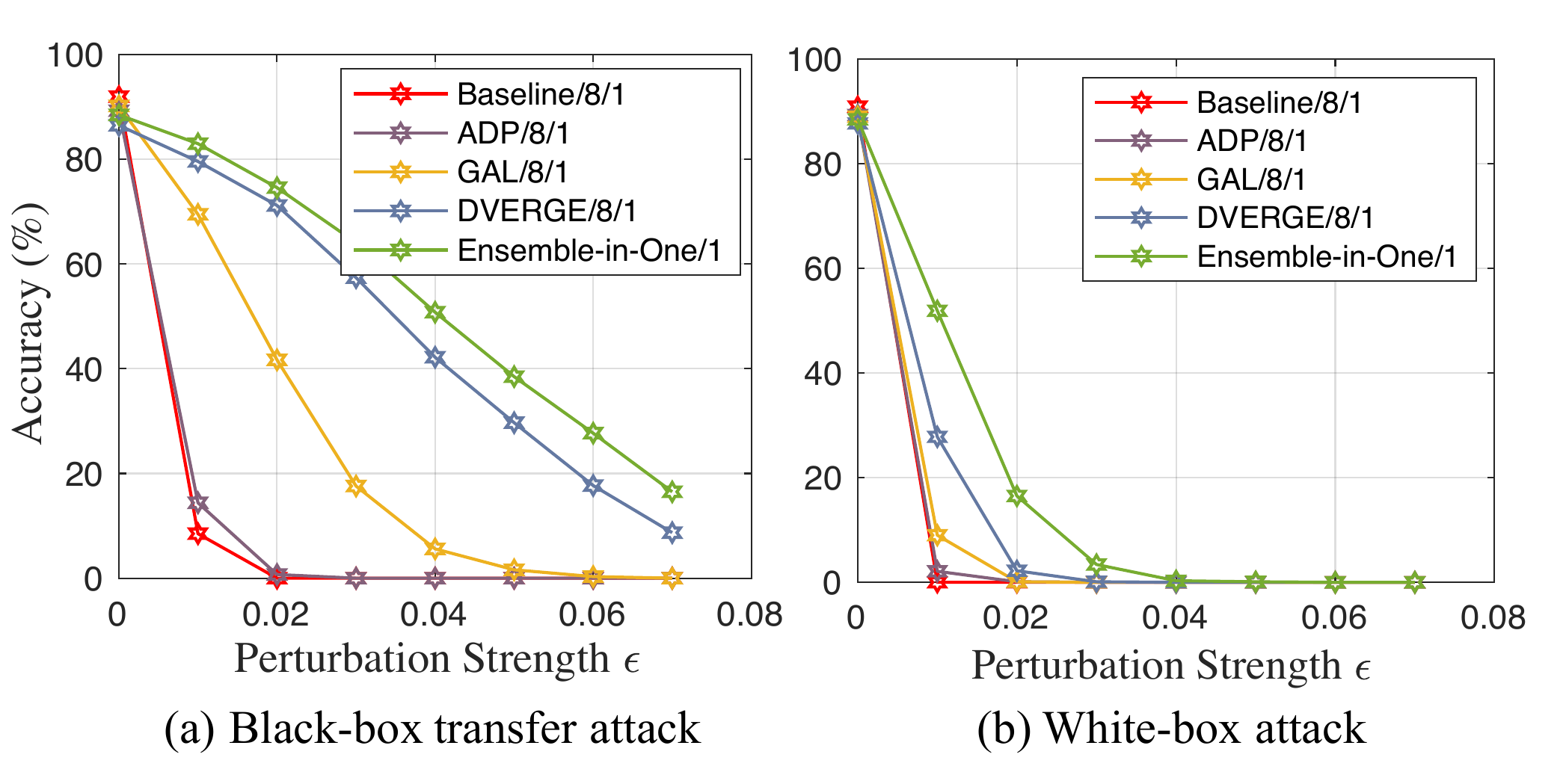}
    \caption{Comparison of the adversarial robustness of the individual models selected from various ensembles. The number after the first slash stands for the number of sub-models within the ensemble, and the number after the second slash means the number of sub-models which are selected to be tested.}
    \label{fig:single_compare}
\end{figure}

\textbf{Comparison of Individual Models.} As illustrated in Sec.\ref{pp:intro}, in real-world application, we prefer deploying more efficient and light models due to the physical hardware constraints and latency requirement. Therefore, we compare the robustness of individual models randomly selected from the ensembles trained by different methods in Fig.\ref{fig:single_compare}. As can be seen, the individual model derived by Ensemble-in-One method consistently outperforms the other individual models selected from the ensembles trained by previous methods. Especially under white-box attack, Ensemble-in-One demonstrates the most remarkable enhancement on the robustness with negligible clean accuracy loss. 

\section{Discussion \& Future Work}
While we have demonstrated and discussed the advantages of Ensemble-in-One, there are also several points that are worthy further exploration. First, the current implementation of augmenting the RGN is simple, by repeating the convolution layers for multiple times. While as observed in Table \ref{tab:n}, enlarging the augmentation factor sometimes brings no benefit on improving the robustness. Hence, there might be better way of constructing the RGN that can compose stronger randomized network, e.g. subtracting some of the unnecessary RGBs. Second, although black-box attacks are more prevalent in real world, defending against white-box attacks is still in demand because recent research warns the high risks of exposing the private models to the adversary \cite{hua2018reverse,hu2020deepsniffer}. Randomized multi-path network can provide promising solutions to addressing the white-box threat concern. If the adversarial transferability among the different paths can be suppressed, the adversarial example generated from one path will be ineffective for another path. Hence, it will make the white-box attacks as difficult as black-box transfer attacks. As also presented in the work mentioned in Sec.\ref{pp:relate_work}, we believe it is a valuable direction to explore defensive method based on randomized multi-path network. 

\section{Conclusions}
In this work, we propose Ensemble-in-One, a novel approach that constructs random gated network (RGN) and learns adversarially robust ensembles within the network. The method is scalable, which can ideally instantiate numerous sub-models by sampling different paths within the RGN. By diversifying the vulnerabilities of different paths, the Ensemble-in-One method can efficiently obtain individual models with higher robustness, simultaneously reducing the overhead of model deployment. The experiments demonstrate the effectiveness of Ensemble-in-One. The individual model derived from the RGN shows much better robustness than the ensembles obtained by previous ensemble training methods.




{\small
\bibliographystyle{ieee_fullname}
\bibliography{egbib}
}

\clearpage
 \onecolumn
\begin{appendices}

 \section{Additional Results}
 In this appendix, we provide some additional results to further compare the advantages and disadvantages of our Ensemble-in-One method and other previous ensemble training methods. 
 
 \subsection{Model Stability Check}
\begin{figure*}[ht]
     \centering
     \hspace{0.1cm}
     \includegraphics[scale=0.48]{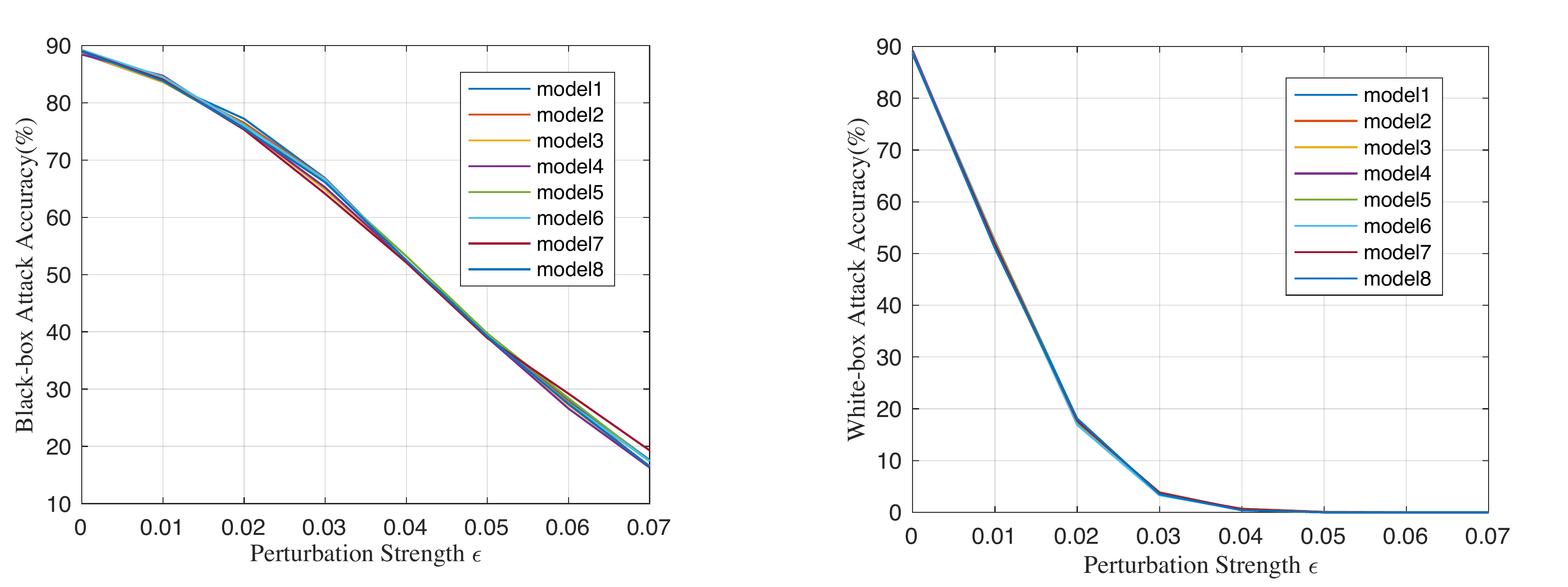}
     \vspace{-0.3cm}
     \caption{The adversarial accuracy versus the perturbation strength against black-box transfer attacks (left) and white-box attacks (right) respectively. Eight different paths are derived from a same random gated network. }
     \label{fig:sblack}
 \end{figure*}

 In the deployment stage, an individual model (or several models) will be derived from the random gated network (RGN) and fine-tuned for a few epochs. Because the model is derived by randomly sampling a path in the RGN, it is important to ensure the stability of derived models. Hence, we randomly derive eight sub-models from a same RGN and test their performance and robustness. As can be observed from Fig.\ref{fig:sblack}, the sampled eight sub-models demonstrate almost the same robustness with very slight fluctuations on the adversarial accuracy against both black-box transfer attacks and white-box attacks. Thus, we confirm that when deriving the sub-models, no additional screening work is required. 

\subsection{Incorporation with adversarial training}
  As similarly done in DVERGE, we augment Ensemble-in-One method with adversarial training (AdvT). Adversarial training can help the models/ensembles obtain better robustness, especially under large perturbation strength and white-box attack scenarios. The underlying reason is that whether DVERGE or our Ensemble-in-One methods, the non-robust features are essentially not eliminated but diversified or shrunken. However, incorporating AdvT will also lead to significant drop on the clean accuracy, because the models will become less sensitive to small changed on the inputs, then for some instances with quite slight difference, the models may not be able to distinguish them.
  
  We integrate the adversarial training with Ensemble-in-One by adding an additional loss, as proposed in DVERGE. Assuming $x_w$ as the adversarial version of $x_s$ which is generated in a white-box manner by utilizing some attack methods (e.g. PGD), the overall optimization goal can be re-written as:
  
  \begin{equation}
     \min \sum_{\forall \mathcal{P}_j \in \mathcal{N}}\mathbb{E}_{(x_t, y_t), (x_s, y_s),l}(\sum_{\forall \mathcal{P}_i \in \mathcal{N}, i\neq j}\mathcal{L}_{f_{\mathcal{P}_j}}(x'_{\mathcal{P}_i^l}(x_t, x_s), y_s) + \mathcal{L}_{f_{\mathcal{P}_j}}(x_w, y_s)).
  \end{equation}
  
\begin{figure*}[h]
      \centering
      \includegraphics[scale=0.5]{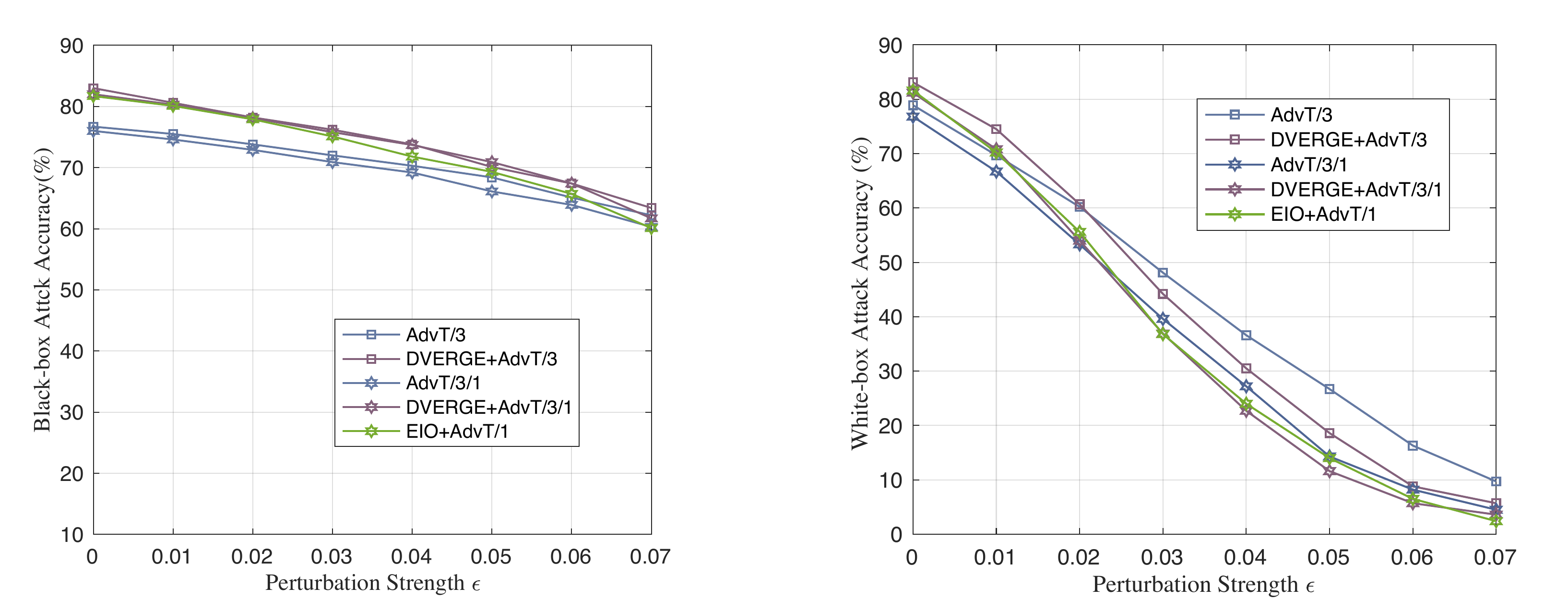}
      \vspace{-0.2cm}
      \caption{The adversarial accuracy versus the perturbation strength against black-box transfer attacks (left) and white-box attacks (right) respectively. For the DVERGE+AdvT and AdvT methods, the number after the first slash represents the number of sub-models contained in the ensemble, and the number after the second slash represents the number of sub-models which are selected from the ensemble for deployment.}
      \label{fig:advt}
    \end{figure*}

  The experimental results show no further improvement than the DVERGE method with adversarial training, as shown in Fig.\ref{fig:advt}. It stands to reason that adversarial training  encourages the models to learn more robust features while leaving less capacity to capture diverse non-robust features. While the basic motivation of Ensemble-in-One is to equivalently instantiate a large number of models to learn from each other. Therefore, the optimization space for Ensemble-in-One will significantly narrowed, thereby only achieving similar performance as DVERGE+AdvT.

  \subsection{Discussion on network augmentation}
  As illustrated in the main manuscript, we augment the original ResNet-20 network to a random gated network (RGN) by augmenting all the convolution layers (in total of 21, each layer is followed by a batchnorm layer) to random gated blocks (RGBs). In fact, it is feasible to flexibly select the augmented layers. As presented in Table \ref{tab:black} and Table \ref{tab:white}, we augment different number of layers in ResNet-20 to construct the RGNs and evaluate their performance. Correspondingly, the distillation layer $l$ for feature distillation will also be bounded, e.g. when only augmenting the top $k$ layers of ResNet-20, the selection of $l$ will be bounded within the range $[1, k]$. 
  
  We find that narrowing the scope of augmented layer can help to improve the clean accuracy, while degrade the adversarial robustness under both black-box and white-box attacks. For example, augmenting \emph{top7} layers of the network obtains a very high clean accuracy. When continuing increasing the augmented layers, the clean accuracy tends to drop while achieving better robustness. These three simple experiments suggest that there are various ways to construct the RGNs and different augmentation tend to capture different performance. Trade-offs between clean accuracy and robustness can be explored by tuning the augmentation. Further exploring better augmentation methods for RGN would also be one of our future goals.

  \begin{table*}[]
      \centering
      \begin{tabular}{c|cccccccc}
      \hline 
         $\epsilon$  & clean & 0.01 & 0.02 & 0.03 & 0.04 & 0.05 & 0.06 & 0.07 \\\hline\hline
         baseline/3/1 & 91.8\% & 7.5\% & 0\% & 0\% & 0\% & 0\% & 0\% & 0\% \\
         baseline/5/1 & 92.2\% & 9.5\% & 0\% & 0\% & 0\% & 0\% & 0\% & 0\% \\
         baseline/8/1 & 92.9\% & 8.3\% & 0\% & 0\% & 0\% & 0\% & 0\% & 0\% \\\hline
         ADP/3/1 & 88.0\% & 18.2\% & 0.7\% & 0\% & 0\% & 0\% & 0\% & 0\% \\
         ADP/5/1 & 90.0\% & 18.5\% & 0.8\% & 0\% & 0\% & 0\% & 0\% & 0\% \\
         ADP/8/1 & 88.7\% & 14.3\% & 0.3\% & 0\% & 0\% & 0\% & 0\% & 0\% \\\hline
         GAL/3/1 & 85.9\% & 71.6\% & 53.8\% & 34.3\% & 18.2\% & 7.7\% & 2.8\% & 0.9\% \\
         GAL/5/1 & 88.9\% & 74.5\% & 52.1\% & 29.6\% & 15.7\% & 6.4\% & 1.9\% & 0.5\% \\
         GAL/8/1 & 89.1\% & 71.0\% & 43.4\% & 20.6\% & 8.2\% & 2.3\% & 0.8\% & 0.4\% \\\hline
         DVERGE/3/1 & 89.5\% & 81.6\% & 67.5\% & 49.6\% & 29.7\% & 15.7\% & 6.3\% & 2.8\% \\
         DVERGE/5/1 & 88.8\% & 81.0\% & 69.2\% & 53.3\% & 37.7\% & 21.9\% & 11.4\% & 3.9\% \\
         DVERGE/8/1 & 86.5\% & 79.6\% & 71.2\% & 57.4\% & 42.2\% & 29.7\% & 17.7\% & 8.7\% \\\hline
         EIO(top7)/1 & 91.2\% & 82.1\% & 71.5\% & 56.6\% & 39.2\% & 25.5\% & 14.6\% & 6.8\% \\
         EIO(top14)/1 & 88.5\% & 82.2\%& 72.5\% & 58.7\% & 44.1\% & 31.7\% & 19.9\% & 12.2\% \\
         EIO(top21)/1 & 88.5\% & 84.0\% & 75.3\% & 64.1\% & 52.1\% & 38.9\% & 29.2\% & 19.3\% \\\hline
         
      \end{tabular}
      \caption{The adversarial accuracy versus the perturbation strength against black-box transfer attacks. We select one of the sub-models within the ensembles which are trained by different methods to test their adversarial accuracy. For our Ensemble-in-One (EIO) method, \emph{topk} means only the top $k$ of the 21 convolution layers are augmented for constructing the random gated network. And the number after the slash means the number of derived models for deployment. For the other methods, the number after the first slash represents the number of sub-models contained in the ensemble, and the number after the second slash represents the number of sub-models which are selected from the ensemble for deployment.}
      \label{tab:black}
  \end{table*}
  
  \begin{table*}[]
      \centering
      \begin{tabular}{c|cccccccc}
      \hline 
         $\epsilon$  & clean & 0.01 & 0.02 & 0.03 & 0.04 & 0.05 & 0.06 & 0.07 \\\hline\hline
         baseline/3/1 & 91.2\% & 0.1\% & 0\% & 0\% & 0\% & 0\% & 0\% & 0\% \\
         baseline/5/1 & 91.7\% & 0.1\% & 0\% & 0\% & 0\% & 0\% & 0\% & 0\% \\
         baseline/8/1 & 90.9\% & 0.1\% & 0\% & 0\% & 0\% & 0\% & 0\% & 0\% \\\hline
         ADP/3/1 & 87.9\% & 3.1\% & 0\% & 0\% & 0\% & 0\% & 0\% & 0\% \\
         ADP/5/1 & 88.9\% & 2.8\% & 0.2\% & 0\% & 0\% & 0\% & 0\% & 0\% \\
         ADP/8/1 & 88.7\% & 2.1\% & 0.1\% & 0\% & 0\% & 0\% & 0\% & 0\% \\\hline
         GAL/3/1 & 86.7\% & 0.3\% & 0.1\% & 0\% & 0\% & 0\% & 0\% & 0\% \\
         GAL/5/1 & 88.2\% & 8.9\% & 0.1\% & 0\% & 0\% & 0\% & 0\% & 0\% \\
         GAL/8/1 & 89.0\% & 9.0\% & 0.1\% & 0\% & 0\% & 0\% & 0\% & 0\% \\\hline
         DVERGE/3/1 & 90.0\% & 13.8\% & 0.2\% & 0\% & 0\% & 0\% & 0\% & 0\% \\
         DVERGE/5/1 & 89.8\% & 20.7\% & 1.3\% & 0.1\% & 0\% & 0\% & 0\% & 0\% \\
         DVERGE/8/1 & 87.7\% & 27.8\% & 2.2\% & 0.1\% & 0\% & 0\% & 0\% & 0\% \\\hline
         EIO(top7)/1 & 91.2\% & 34.1\% & 4.3\% & 0.3\% & 0\% & 0\% & 0\% & 0\% \\
         EIO(top14)/1 & 88.5\% & 41.4\%& 9.5\% & 0.7\% & 0.1\% & 0\% & 0\% & 0\% \\
         EIO(top21)/1 & 89.0\% & 52.4\% & 18.0\% & 3.4\% & 0.6\% & 0\% & 0\% & 0\% \\\hline
         
      \end{tabular}
      \caption{The adversarial accuracy versus the perturbation strength against black-box transfer attacks. We select one of the sub-models within the ensembles which are trained by different methods to test their adversarial accuracy. The notations are the same as Table \ref{tab:black}. The clean accuracy is slightly different with Table \ref{tab:black} because the instances used for evaluating black-box and white-box attacks are from two groups of randomly sampled images. We test the accuracy against black-box attack on the same set of adversarial examples as DVERGE, while sampling another set of data to test the accuracy against white-box attacks because the random seed changes. }
      \label{tab:white}
  \end{table*}
  
\end{appendices}

\end{document}